\pdfoutput = 1
\documentclass[final]{article}

\PassOptionsToPackage{numbers, compress}{natbib}
\usepackage[]{nips_2016}

\usepackage[utf8]{inputenc} 
\usepackage[T1]{fontenc}    
\usepackage{url}            
\usepackage{booktabs}       
\usepackage{amsfonts}       
\usepackage{amsmath}
\usepackage{nicefrac}       
\usepackage{microtype}      
\usepackage{xcolor}
\usepackage{graphicx}
\usepackage{wrapfig}

\DeclareMathOperator*{\argmax}{arg\max}
\newcommand{\BO}{GPBO}
\title{Hybrid Repeat/Multi-point Sampling for Highly Volatile Objective Functions}

\author{
  Brett W. Israelsen\thanks{Graduate Researcher, corresponding author, \url{http://bisraelsen.github.io/}} \\
  Department of Computer Science\\
  University of Colorado\\
  Boulder, CO 80309\\
  \texttt{brett.israelsen@colorado.edu} \\
  \And
  Nisar ~Ahmed\thanks{Assistant Professor, \url{http://www.cohrint.info/}} \\
  Department of Aerospace Engineering Sciences\\
  University of Colorado\\
  Boulder, CO 80309\\
  \texttt{nisar.ahmed@colorado.edu} \\
}

\begin{document}

\maketitle
\begin{abstract}
    A key drawback of the current generation of artificial decision-makers is that they do not adapt well to changes in unexpected situations.  This paper addresses the situation in which an AI for aerial dog fighting, with tunable parameters that govern its behavior, will optimize behavior with respect to an objective function that must be evaluated and learned through simulations. Once this objective function has been modeled, the agent can then choose its desired behavior in different situations. Bayesian optimization with a Gaussian Process surrogate is used as the method for investigating the objective function. One key benefit is that during optimization the Gaussian Process learns a global estimate of the true objective function, with predicted outcomes and a statistical measure of confidence in areas that haven't been investigated yet. However, standard Bayesian optimization does not perform consistently or provide an accurate Gaussian Process surrogate function for highly volatile objective functions. We treat these problems by introducing a novel sampling technique called Hybrid Repeat/Multi-point Sampling. This technique gives the AI ability to learn optimum behaviors in a highly uncertain environment. More importantly, it not only improves the reliability of the optimization, but also creates a better model of the entire objective surface. With this improved model the agent is equipped to better adapt behaviors.
\end{abstract}

\section{Introduction}
    Due to current and expected logistical and fiscal constraints the Department of Defense (DoD) has been focusing on simulation-based training of warfighters. To this end, the Not-So-Grand Challenge was developed, with the specific goal to investigate solutions for current and future simulation training systems. As part of this challenge different autonomous agents were developed and evaluated based on their ability to mimic a human pilot in given situations~\cite{Doyle2014}. Even though an autonomous agent may mimic a human pilot there still remains the question of whether it can adapt based on the adversaries responses.
    
    Previous related work focused on optimization of target allocation, tactics, and mission plans for aerial combat~\cite{Mulgund1998, Mulgund2001,Wu2005,Gonsalves2004}, but have not addressed adaptation of autonomous AI decision-makers with tunable behavioral parameters.  This work specifically examines how an agent with tunable parameters that govern overall behavior can be adapted to optimize an objective function that quantifies engagement outcomes. Beyond optimizing some outcome metric, it is also important that the agent have a realistic representation of the entire objective function. This will allow the AI to anticipate the likelihood of successful engagements with adversaries (human or AI) under different uncertain conditions. These behavioral changes could be based on adapting to the adversary's skill level (it is not desirable to use the same difficulty level for novice and advanced pilots), or the adversary's ability to exploit a weakness of the agent (the objective function is changing).

    There are several challenges that make optimization of AI behavioral parameters difficult in this application:
    
    \begin{enumerate}
        \item Simulating an engagement can be costly. Beyond the financial expense of operating the simulation environment, contributions to the cost may also include the involvement of skilled labor/participants with limited availability, and the wall-clock duration of the simulation itself.\label{pnt:cost}
        \item The objective function to be optimized is not known a priori, and when sampled is generally nonlinear and noisy. Consequently, many traditional optimization methods are not applicable.\label{pnt:uglyobjective}
        \item Due to the realistic nature of the simulations and the nature of aerial combat, the objective function is extremely volatile and uncertain (e.g. due to combined random effects of weather, terrain, sensor noise, psycho-motor time delays, etc.).\label{pnt:volatile}
        \item Besides only identifying the optimum performance, the agent should also try to obtain some model of the overall objective function. This can allow the agent to be adaptive and have some notion of what outcomes might arise when modifying behavior parameters, without having to exhaustively search over a high-dimensional parameter space. In addition,  this model can also be used to generate a useful estimate of the expected performance of adversaries for a wide range of scenarios, using only a small number of test evaluations. \label{pnt:objectivemodel}
    \end{enumerate}
    
    Gaussian Process based Bayesian optimization (\BO{}) is well-suited for addressing points~\ref{pnt:cost} and~\ref{pnt:uglyobjective}. However, we show that typical application of \BO{}~is not well suited to address points~\ref{pnt:volatile} and~\ref{pnt:objectivemodel}, in this application. We introduce and demonstrate a new sampling approach called Hybrid Repeat/Multi-point Sampling (HRMS) that yields some promising results in this regard, by capturing more statistical information about the objective function on each iteration of the optimization. With proper configuration, HRMS is able to not only identify the optima more reliably than standard \BO{}, but also yields a more useful surrogate representation of the objective surface. Finally, it generally does this using no more total function evaluations that traditional \BO.

\section{Methodology}
    \begin{wrapfigure}{R}{0.48\textwidth}
        \centering
        \includegraphics[width=0.5\textwidth]{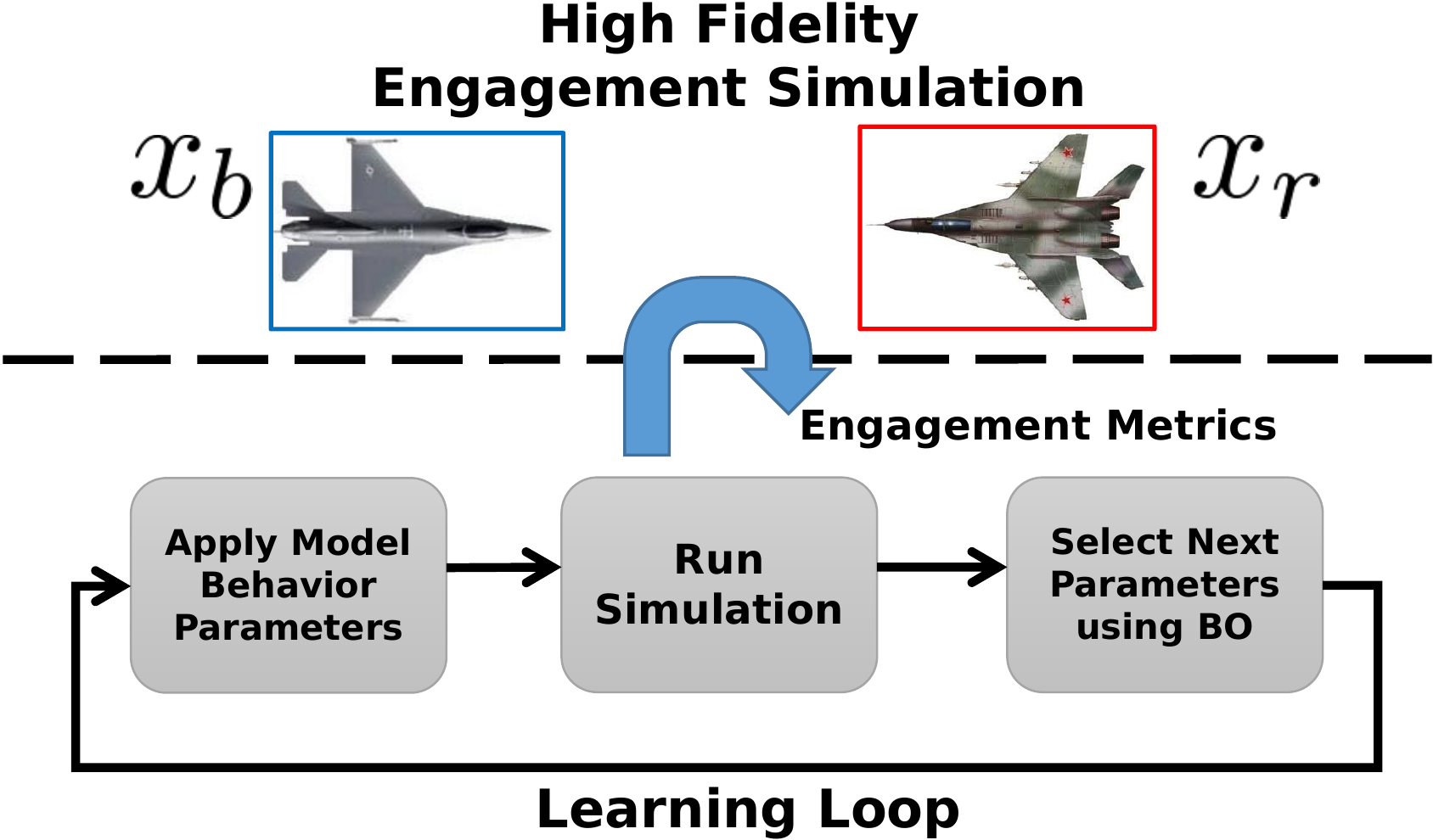}
        \caption{Learning Process Diagram. Representation of the engagement simulation environment (top) and the high-level learning loop (bottom)}
        \label{fig:LearningEngine}
    \end{wrapfigure}
     The problem is defined as an air combat scenario with autonomous red and blue force agents. Each of the agents has behavioral parameters given by the parameter vectors $\pmb{x}_{r}$ and $\pmb{x}_{b}$ respectively. The goal is to optimize an objective function $y_i(\pmb{x}_{r},\pmb{x}_{b})$, where $y_i(\cdot)$ must be evaluated using a high-fidelity combat simulation. For this work $\pmb{x}_r$ is constant, and the optimization will only be changing $\pmb{x}_b$. The remainder of this paper uses the following: $y=y_{TTK}(\cdot)$ (TTK stands for time to kill), and $\pmb{x_b} = \{x_1,x_2\} = \{launch,intspeed\}$, where  $launch$ is the time to launch weapon once lock is acquired and $intspeed$ is the intercept speed (i.e. the speed at which the blue fighter moves into close the range on red once engaged). Note that there are 11 total behavior parameters available in the aerial-combat simulation, but we only investigate two here.

\begin{wrapfigure}{R}{0.55\textwidth}
        \centering
        \includegraphics[height=3.0cm]{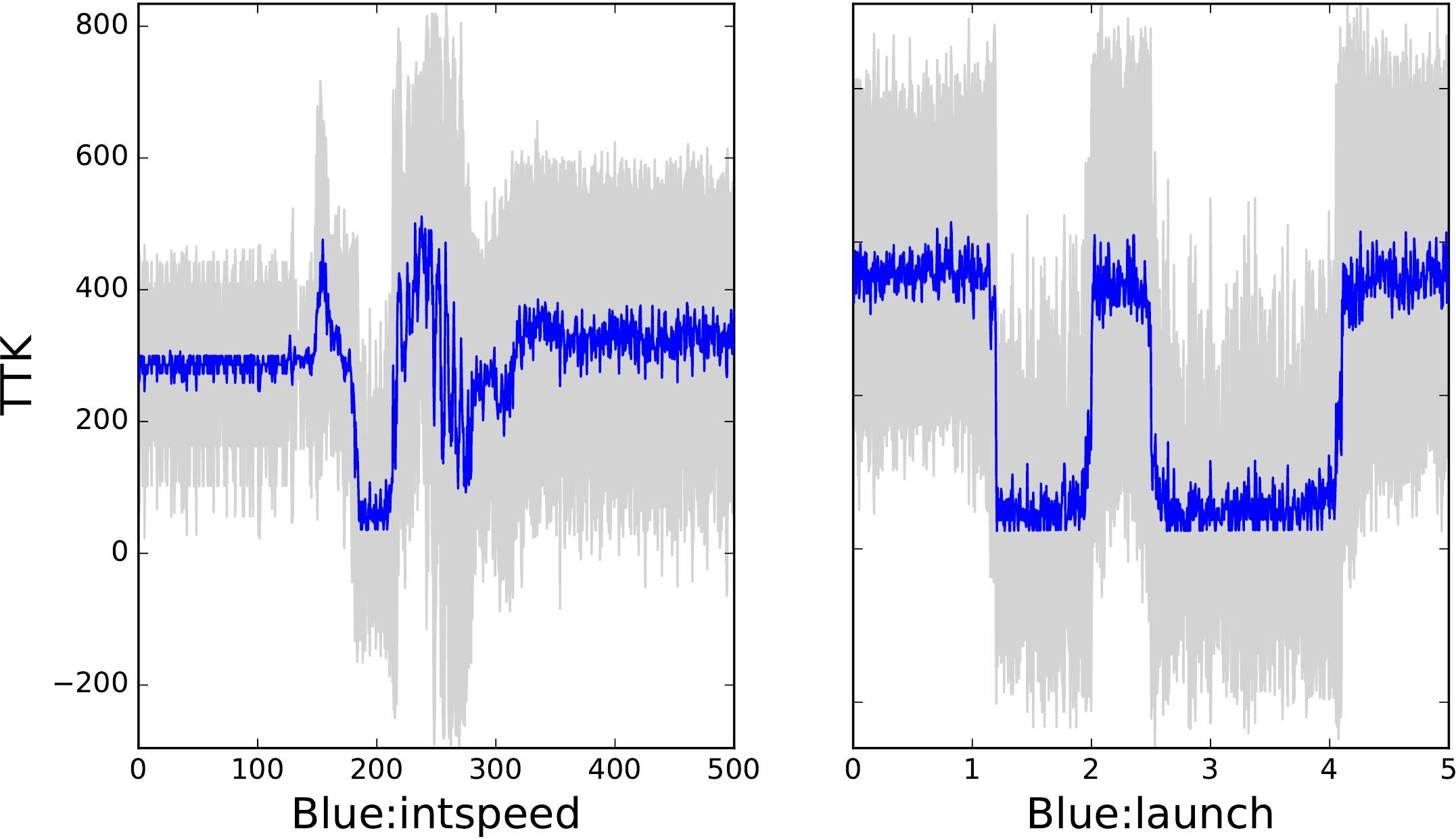}
        \caption{One-dimensional examples of \emph{TTK} objective function, for $\pmb{x}=\{\emph{Blue:intspeed}, \emph{Blue:launch}\}$. These figures were produced by holding all $\pmb{\pmb{x}_r}$ parameters constant, as well as all $\pmb{x}_b$ parameters except the one listed. The dark blue line represents the empirical $\mu$ and the shaded area is $2\sigma$}
        \label{fig:objective_examples}
    \end{wrapfigure}

    Figure~\ref{fig:LearningEngine} depicts the high-level learning process. Both the red and blue agents have tunable parameters, but only blue is being changed in this scenario. Figure~\ref{fig:objective_examples} shows 1d slices of the objective function for $x_b=intspeed$ and $x_b=launch$.

    Using an acquisition function $a(\cdot)$, there are two commonly used ways to search for the optimum. The first is single sampling (SS) where $y_{i}(\pmb{x})$ is evaluated a single time at the $\argmax_{\pmb{x}}a(\cdot)$ of the acquisition function; this is the standard \BO{} approach. The second is multiple (or batch) sampling (MS), in which $y_{i}(\pmb{x})$ is evaluated at multiple different locations simultaneously. Finally, we propose a method called repeat sampling (RS), which is identical to SS except that the objective function will be evaluated repeatedly at the same location, $\argmax_{\pmb{x}}a(\cdot)$.
    
    The intuitive reason for introducing repeat sampling is to obtain a more informative statistical sample of the objective function at every iteration. This is necessary because, for \BO{} to work properly, the surrogate function needs to be a `sufficiently accurate' representation of the true objective function. RS helps the GP to have more information regarding the noise of the true objective function, so that the GP can be a useful surrogate function in guiding the \BO. This method is also used in traditional experimental design where it is called `replication'(see \cite{pyzdek2003quality} and \cite[sec. 4.4.4.6]{croarkin2006nist}). Another consideration that makes repeat sampling or replication attractive is prohibitive cost to setting up new experiments, something that is not as much of a problem when dealing with computer simulations but becomes an important consideration with applied problems like training pilots.

    The RS and MS strategies would be especially valuable when the objective function is less expensive to evaluate via simulation, and the experiments can be run in parallel without significantly increasing the overall cost of the optimization. In the following, MS=3 means that 3 batch samples will be selected. Likewise, RS=3 is where 3 samples will be taken at the same location. Note that SS is a special case where RS=MS=1. Finally, we refer to combined sampling where RS and MS are both greater than 1, as Hybrid Repeat/Multi-point Sampling (HRMS).

\section{Experiments}
    Given a decision agent optimization problem, we perform experiments to investigate the performance of different acquisition functions for the aerial combat simulations. More importantly we wish to investigate the effect that varying RS and MS has on the optimization results.

    Specifically, we evaluate three different, common, acquisition functions: Expected Improvement (EI)~\cite{Jones1998}, upper confidence bound (GP-UCB)~\cite{Srinivas2010}, and Thompson Sampling (TS)~\cite{Thompson1933}. We also evaluate their corresponding batch sampling forms: q-EI~\cite{Wang2016}, GP-UCB-PE ~\cite{Contal2013}, and multiple draws from TS. The different levels of RS and MS used are $RS=\{1,3,5,10\}$ and $MS=\{1,3,5\}$. The GPML toolbox~\cite{Rasmussen2006b} is used for GP representation and hyperparameter inference. There is also an interface to the MSS air combat simulation engine made by Orbit Logic Inc. The kernel is the M\'{a}tern ARD kernel with $\nu=3/2$~\cite{Rasmussen2006}. \BO{} is run for approximately 500 function evaluations (approximately 500 because different HRMS configurations don't allow exactly 500); 4 experiments are run per HRMS configuration using 20 random seed locations to bootstrap GP learning with MAP inference for hyperparameter optimization.

\begin{figure}[htbp]
        \centering
        \includegraphics[width=0.90\textwidth]{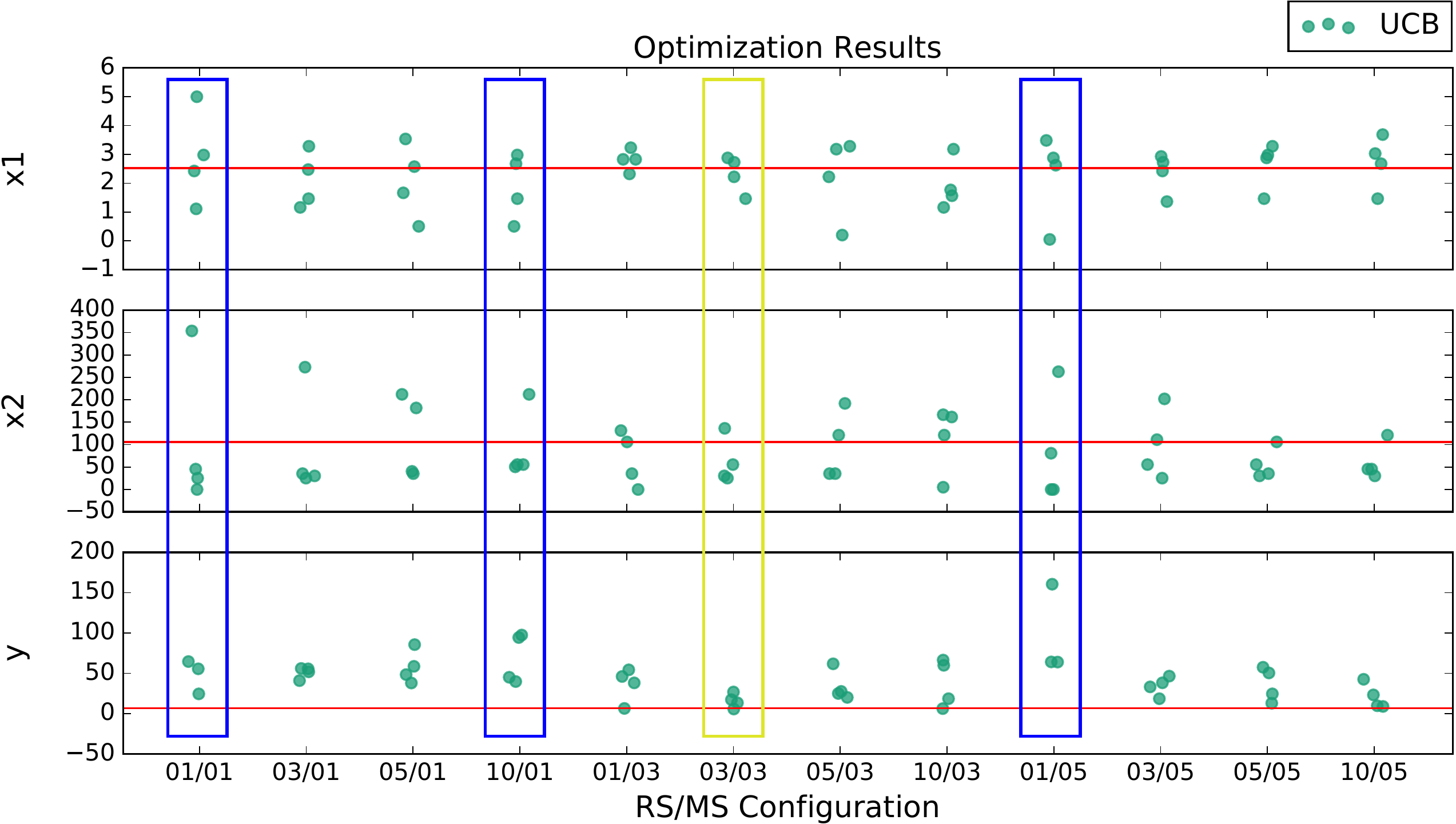}
        \caption{Scatter plots of the $x1,x2$ and $y$ values for different RS/MS configurations. Results from running \BO{} for approximately 500 function evaluations. The red horizontal line is the ground truth value.}
        \label{fig:2d_time_plot_comparison}
    \end{figure}

    Figure~\ref{fig:2d_time_plot_comparison} shows the estimated locations of the optimum as well as the optimum itself for the UCB acquisition function. The estimates for $\pmb{x}$ and $y$ grow tighter together, and closer to the ground truth, as both RS and MS become greater than 1. The configurations marked by colored rectangles highlight that methods using solely SS, RS, and MS (shown in blue), underperform the method that combines both RS and MS greater than one (yellow). This finding is similar for the EI and TS functions as well. From this figure we can conclude that there are HRMS configurations that yield more repeatable optimization results than SS, and that neither RS or MS alone is clearly better.

    On more detailed investigation of the results, for large RS the optimization tends to terminate early due to an ill-conditioned covariance matrix. This occurred because RS is `too big' at those locations, having returned too many nearly identical objective function values at the same $(x_1,x_2)$ locations. The subsequent covariance matrix became too linearly dependent, which led to conditioning problems for GP inference. This suggests that there is clearly a trade-off between the benefits of RS and an unstable GP. We revisit this point in the conclusion section.

\begin{figure}[htbp]
        \centering
        \includegraphics[width=0.90\textwidth]{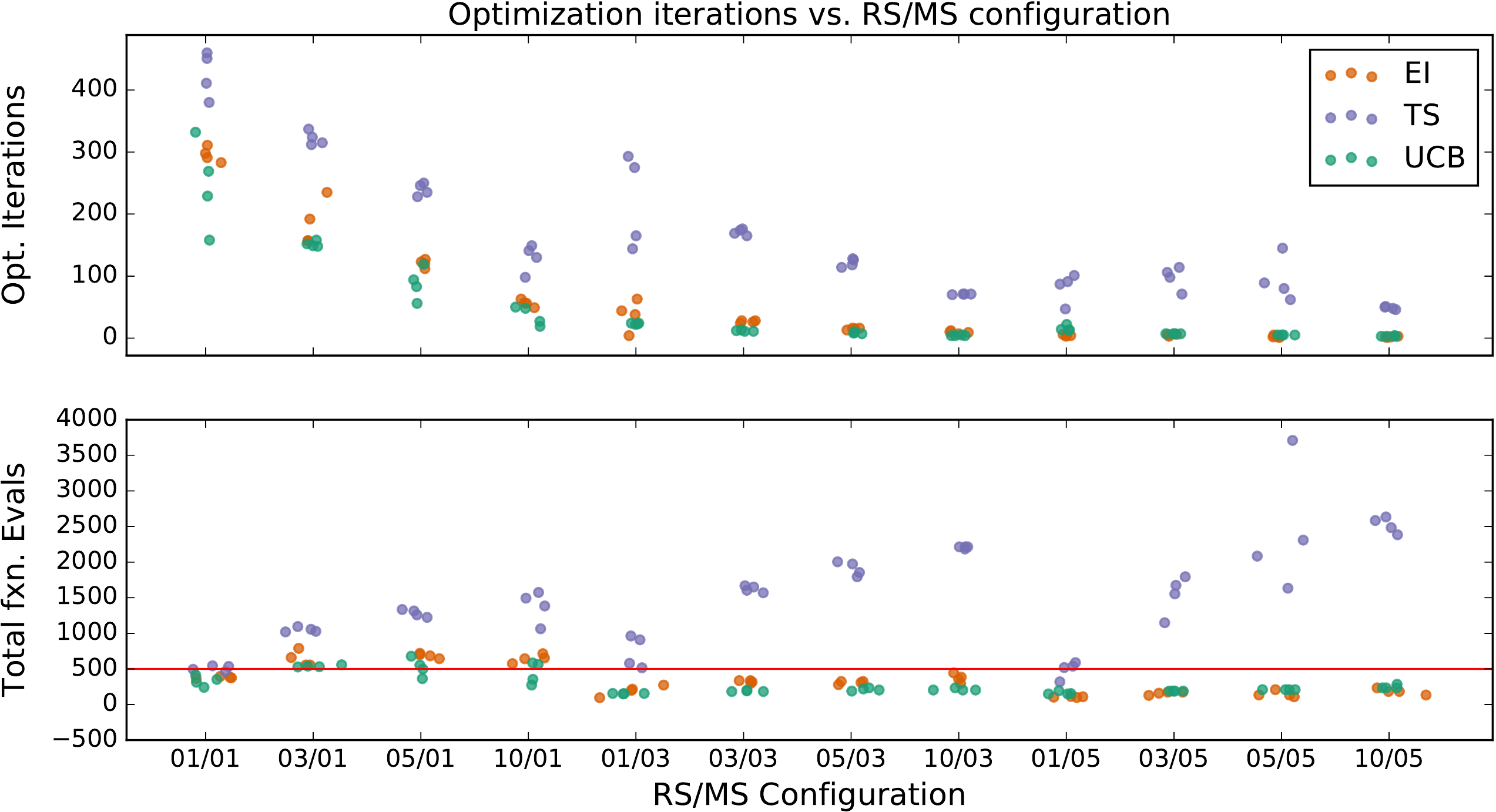}
        \caption{ Plot showing the total amount of optimization iterations (Top), and the corresponding number of function evaluations (Bottom) after running each configuration for 2 hours. Note that, with the exception of TS, the total number of function evaluations for mixed RS/MS configurations generally doesn't exceed that of \BO{} with SS}
        \label{fig:tot_evaluations}
    \end{figure}

    It is important to note that given a fixed time for optimization, in other words: not limiting the function evaluations for methods that perform more quickly, the total number of function evaluations in most cases does not exceed that of the SS strategy. This is mainly due to the overhead of calculating MS, and is illustrated in Figure~\ref{fig:tot_evaluations}, where only TS significantly exceeds the total function evaluations of SS. 

\begin{figure}[htbp]
        \centering
        \includegraphics[width=0.90\textwidth]{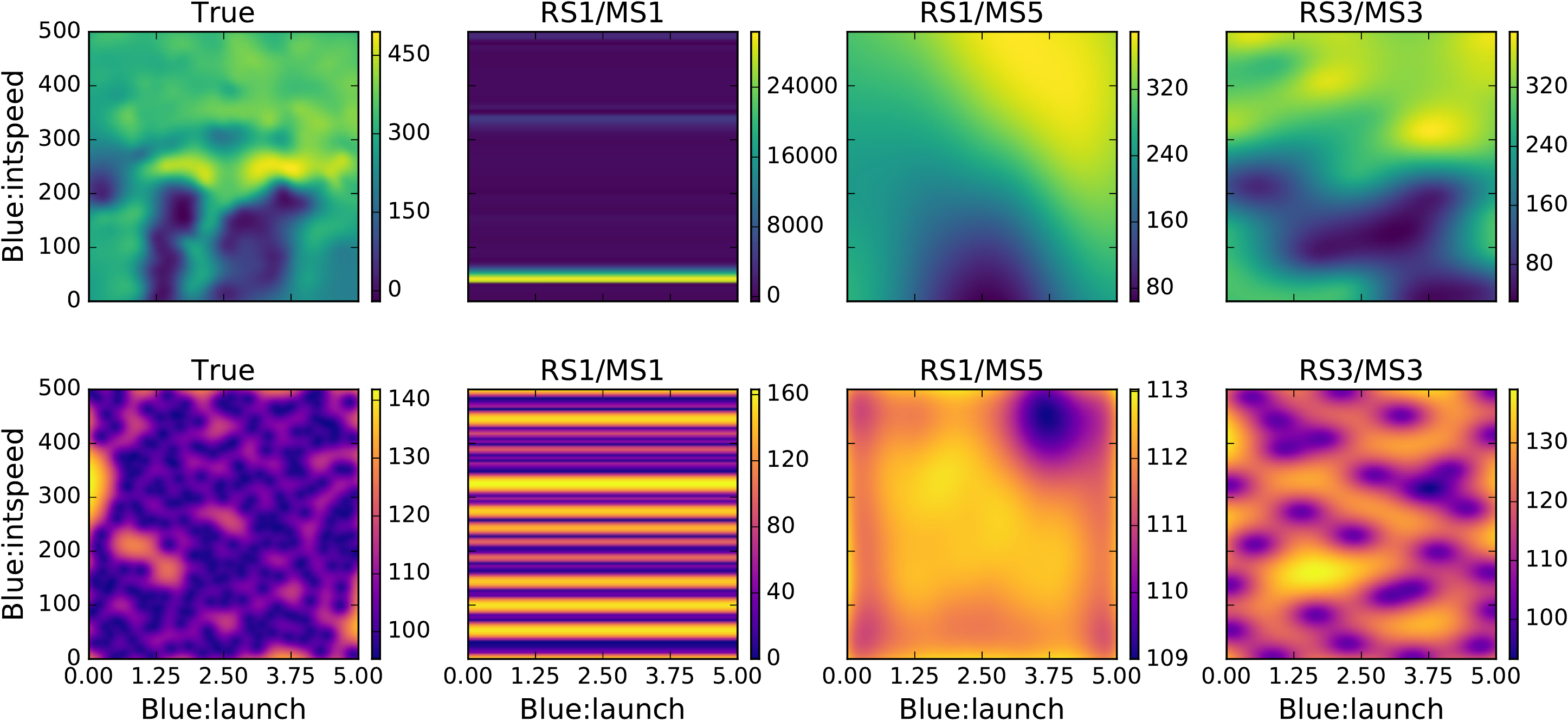}
        \caption{Table of figures illustrating the effect of combined RS/MS sampling using the UCB acquisition function. Top row is $\mu_{GP}$ bottom row is $\sigma_{GP}$. From left to right the first column is the truth surface obtained by high density sampling and fitting a GP to the data. The following columns show some example results from optimization runs using the indicated values for RS and MS. Each of the final 3 columns represents the optimization solution after 2 hours}
        \label{fig:sample_instability}
    \end{figure}

    Figure~\ref{fig:sample_instability} depicts some examples of the final GPs obtained during time limited optimization (again, allowing faster methods to use more evaluations/iterations) for three different HRMS configurations. The far left column is the `ground truth GP' model that is obtained by training with several thousands of samples over the input space. The key insight is that the RS3/MS3 strategy yielded a GP that better represents the underlying stochastic function. 
    
    These findings indicate that HRMS both improves the repeatability of the optimization, \emph{and} the overall fidelity of the surrogate representation of the objective function. This applies for both a fixed computation time (i.e. not limiting faster methods like SS to have the same number of function evaluations), and number of function evaluations. These two phenomena are linked, i.e. the optimization is more repeatable (and reliable) because the surrogate representation is more accurate.

\section{Conclusion}
    We have shown promising preliminary results of HRMS, a novel sampling strategy that helps improve both the repeatability/reliability of \BO{} (a desirable feature for box optimization), and a better surrogate representation of the true objective surface. This surface can be used in understanding how the underlying process works and will allow an AI decision-maker to adapt in highly volatile and uncertain environments. Preliminary experiments show that improvements from HRMS are independent of the acquisition function for our application. They also show that the improved performance is due to the fact that adding both local and global information about the objective function at each time step makes the surrogate function more accurate. This accuracy yields a more efficient \BO{} process, and does not require more function evaluations than the standard SS approach. Again, these results while promising still need to be examined more rigorously and be statistically verified. 
    
    To the best of our knowledge, HRMS has not been considered for GPBO previously.  Further investigation regarding the relationship with replication, in design of experiments, needs to be explored more formally. There is still much work to be done regarding automatic selection of RS and MS. Perhaps only adding new data to the GP covariance function if it has sufficient variation might work, but would need to be formally assessed. Finally, the preliminary results shown here are being extended and verified in higher dimensional problems. The AI decision-maker in our aerial combat simulation, for instance, has 11 behavioral parameters that can be used for optimization. In higher dimensional spaces, the automatic selection criteria for RS and MS becomes more important as the surrogate function and optimization results become much more unwieldy and uncertain, and visual comparison is no longer available to verify the similarity between the true objective function and its surrogate representation.

\newpage
\subsubsection*{Acknowledgments}
We would like to acknowledge Kenneth Center, and Roderick Green with Orbit Logic Incorporated for their collaboration in designing the simulation framework used.
\bibliographystyle{unsrt}
\bibliography{References.bib}

\begin{thebibliography}{10}

\bibitem{Doyle2014}
Margery~J Doyle and Antoinette~M Portrey.
\newblock {Rapid Adaptive Realistic Behavior Modeling is Viable for Use in
  Training}.
\newblock In {\em Proceedings of the 23rd Conference on Behavior Representation
  in Modeling and Simulation (BRIMS)}, 2014.

\bibitem{Mulgund1998}
S.~Mulgund, K.~Harper, K.~Krishnakumar, and G.~Zacharias.
\newblock {Air combat tactics optimization using stochastic genetic
  algorithms}.
\newblock In {\em SMC'98 Conference Proceedings. 1998 IEEE International
  Conference on Systems, Man, and Cybernetics (Cat. No.98CH36218)}, volume~4,
  pages 3136--3141, Oct 1998.

\bibitem{Mulgund2001}
Sandeep Mulgund, Karen Harper, and Greg Zacharias.
\newblock {Large-Scale Air Combat Tactics Optimization Using Genetic
  Algorithms}.
\newblock {\em Journal of Guidance, Control, and Dynamics}, 24(1):140--142, Jan
  2001.

\bibitem{Wu2005}
Wen-Hai Wu and Qingdao Branch.
\newblock {Air Combat Decision-making for Cooperative Multiple Target Attack
  using Heuristic Adaptive Genetic Algorithm}.
\newblock In {\em Proceedings of the 4th International Conference on Machine
  Learning and Cybernetics}, volume~1, pages 473--478, Aug 2005.

\bibitem{Gonsalves2004}
P.G. Gonsalves and J.E. Burge.
\newblock {Software Toolkit for Optimizing Mission Plans (STOMP)}.
\newblock In {\em Collection of Technical Papers - AIAA 1st Intelligent Systems
  Technical Conference}, volume~1, pages 391--399. American Institute of
  Aeronautics and Astronautics, Sep 2004.

\bibitem{pyzdek2003quality}
Thomas Pyzdek and Paul~A Keller.
\newblock {\em Quality engineering handbook}.
\newblock CRC Press, 2003.

\bibitem{croarkin2006nist}
Carroll Croarkin and Paul Tobian.
\newblock {e-Handbook of Statistical Methods}.
\newblock {\em NIST/SEMATECH, Available online: http://www. itl. nist.
  gov/div898/handbook}, 2016.

\bibitem{Jones1998}
Donald~R. Jones, Matthias Schonlau, and J~William.
\newblock {Efficient Global Optimization of Expensive Black-Box Functions}.
\newblock {\em Journal of Global Optimization}, 13(4):455--492, 1998.

\bibitem{Srinivas2010}
Niranjan Srinivas, Andreas Krause, Sham~M. Kakade, and Matthias Seeger.
\newblock {Gaussian Process Optimization in the Bandit Setting: No Regret and
  Experimental Design}.
\newblock Jun 2010.

\bibitem{Thompson1933}
William~R. Thompson.
\newblock {On the Likelihood that One Unknown Probability Exceeds Another in
  View of the Evidence of Two Samples}.
\newblock {\em Biometrika}, 25(3/4):285, Dec 1933.

\bibitem{Wang2016}
Jialei Wang, Scott~C. Clark, Eric Liu, and Peter~I. Frazier.
\newblock {Parallel Bayesian Global Optimization of Expensive Functions}.
\newblock Feb 2016.

\bibitem{Contal2013}
Emile Contal, David Buffoni, Alexandre Robicquet, and Nicolas Vayatis.
\newblock {Parallel Gaussian process optimization with upper confidence bound
  and pure exploration}.
\newblock In {\em Lecture Notes in Computer Science (including subseries
  Lecture Notes in Artificial Intelligence and Lecture Notes in
  Bioinformatics)}, volume 8188 LNAI, pages 225--240. Springer, 2013.

\bibitem{Rasmussen2006b}
Carl~Edward Rasmussen, Christopher K~I Williams, and ebrary Inc.
\newblock {GPML Toolbox}, Dec 2006.

\bibitem{Rasmussen2006}
Carl~Edward Rasmussen and Christopher K.~I. Williams.
\newblock {\em {Gaussian processes for machine learning}}.
\newblock Adaptive computation and machine learning. MIT Press, Cambridge,
  Mass, 2006.

\end{thebibliography}

\end{document}